\documentclass[10pt,twocolumn,letterpaper]{article}
\usepackage{cvpr}
\usepackage{times}
\usepackage{epsfig}
\usepackage{graphicx}
\graphicspath{{latex/Figures/}}
\graphicspath{{Figures/}}
\usepackage{amsmath,amsfonts, amssymb}
\usepackage{relsize}
\usepackage{tabularx}
\newcolumntype{Y}{>{\centering\arraybackslash}X}
\usepackage{multirow}
\usepackage{blindtext}
\usepackage{booktabs}
\usepackage{bm}
\usepackage{mmstyles}
\usepackage[section]{placeins}
\usepackage[numbers,sort]{natbib}
\usepackage{diagbox}
\usepackage{caption}
\usepackage[percent]{overpic}
\usepackage{enumitem}

\usepackage{cuted}
\usepackage{capt-of}

\usepackage[dvipsnames]{xcolor}
\definecolor{todo}{rgb}{1,.5,0} 

\usepackage[pagebackref=true,breaklinks=true,letterpaper=true,colorlinks,bookmarks=false]{hyperref}
\usepackage[british,UKenglish,USenglish,english,american]{babel}

\cvprfinalcopy 

\def\cvprPaperID{3033} 

\ifcvprfinal\fi

\begin{document}

\title{Bringing Old Photos Back to Life}

\author{Ziyu Wan$^{1*}$, Bo Zhang$^2$, Dongdong Chen$^3$, Pan Zhang$^4$, Dong Chen$^2$, Jing Liao$^{1\dagger}$, Fang Wen$^2$ \\
	$^1$City University of Hong Kong \quad  
	$^2$Microsoft Research Asia \quad $^3$Microsoft Cloud + AI\\
	$^4$University of Science and Technology of China
	}

\twocolumn[{
\renewcommand\twocolumn[1][]{#1}
\maketitle
    \vspace{-1.5em}
    \setlength\tabcolsep{0.5pt}
    \centering
    \small
    \begin{tabular}{c}
        \includegraphics[width=0.99\textwidth]{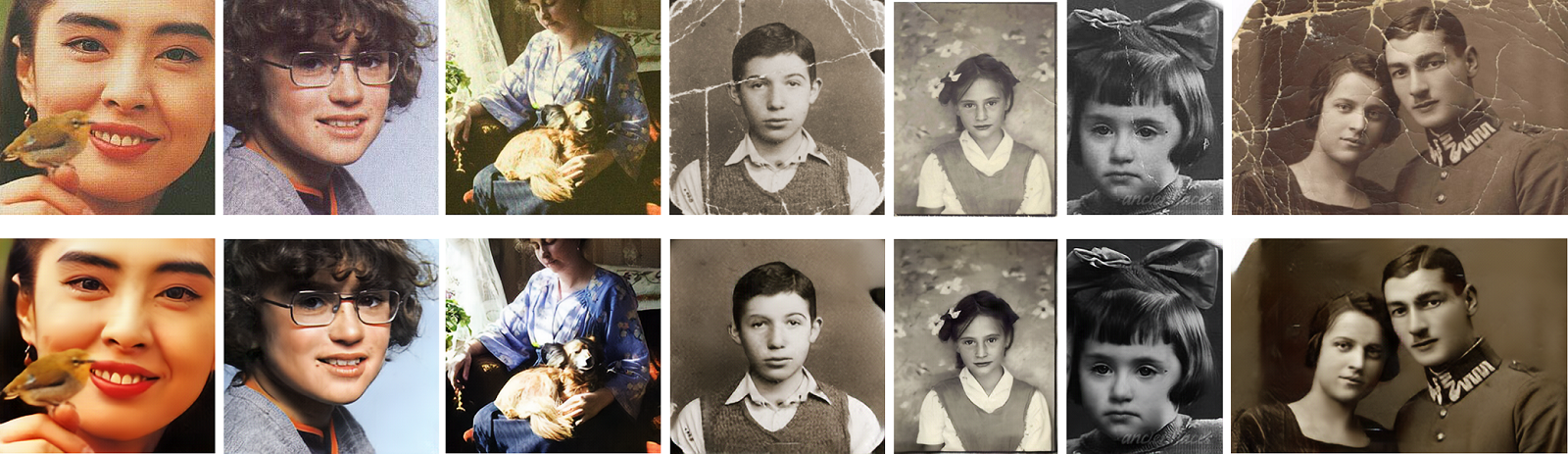}
    \end{tabular}
    \vspace{-1em}
    \captionof{figure}{\textbf{Old image restoration results produced by our method.} Our method can handle the complex degradation mixed by both unstructured and structured defects in real old photos. \textit{Project Website:} \url{http://raywzy.com/Old_Photo/}}
    \label{fig:teaser}
    \vspace{1em}
}]

\maketitle

\begin{abstract}

    \vspace{-0.9em}
    We propose to restore old photos that suffer from severe degradation through a deep learning approach. Unlike conventional restoration tasks that can be solved through supervised learning, the degradation in real photos is complex and the domain gap between synthetic images and real old photos makes the network fail to generalize. Therefore, we propose a novel triplet domain translation network by leveraging real photos along with massive synthetic image pairs. Specifically, we train two variational autoencoders (VAEs) to respectively transform old photos and clean photos into two latent spaces. And the translation between these two latent spaces is learned with synthetic paired data. This translation generalizes well to real photos because the domain gap is closed in the compact latent space. Besides, to address multiple degradations mixed in one old photo, we design a global branch with a partial nonlocal block targeting to the structured defects, such as scratches and dust spots, and a local branch targeting to the unstructured defects, such as noises and blurriness. Two branches are fused in the latent space, leading to improved capability to restore old photos from multiple defects. The proposed method outperforms state-of-the-art methods in terms of visual quality for old photos restoration.{\let\thefootnote\relax\footnotetext{$^{*}$ Work done during the internship at Microsoft Research Asia}} {\let\thefootnote\relax\footnotetext{$^{\dagger}$ Corresponding author}} 
\end{abstract}

\section{Introduction}
\label{sec:introuction} 
    Photos are taken to freeze the happy moments that otherwise gone. Even though time goes by, one can still evoke memories of the past by viewing them. Nonetheless, old photo prints deteriorate when kept in poor environmental condition, which causes the valuable photo content permanently damaged. Fortunately, as mobile cameras and scanners become more accessible, people can now digitalize the photos and invite a skilled specialist for restoration. However, manual retouching is usually laborious and time-consuming, which leaves piles of old photos impossible to get restored. Hence, it is appealing to design automatic algorithms that can instantly repair old photos for those who wish to bring old photos back to life.

    Prior to the deep learning era, there are some attempts~\cite{stanco2003towards,bruni2004generalized,chang2005photo,giakoumis2005digital} that restore photos by automatically detecting the localized defects such as scratches and blemishes, and filling in the damaged areas with inpainting techniques. Yet these methods focus on completing the missing content and none of them can repair the spatially-uniform defects such as film grain, sepia effect, color fading, etc., so the photos after restoration still appear outdated compared to modern photographic images. With the emergence of deep learning, one can address a variety of low-level image restoration problems~\cite{zhang2017learning,zhang2017beyond,dong2014learning,xu2014deep,ren2016single,zhang2019deep,he2018deep,chen2018gated} by exploiting the powerful representation capability of convolutional neural networks, \ie, learning the mapping for a specific task from a large amount of synthetic images. 
    
    The same framework, however, does not apply to old photo restoration. First, the degradation process of old photos is rather complex, and there exists no degradation model that can realistically render the old photo artifact. Therefore, the model learned from those synthetic data generalizes poorly on real photos. Second, old photos are plagued with a compound of degradations and inherently requires different strategies for repair: unstructured defects that are spatially homogeneous, \eg, film grain and color fading, should be restored by utilizing the pixels in the neighborhood, whereas the structured defects, \eg, scratches, dust spots, etc., should be repaired with a global image context.

    To circumvent these issues, we formulate the old photo restoration as a triplet domain translation problem. Different from previous image translation methods~\cite{isola2017image}, we leverage data from three domains (\ie, real old photos, synthetic images and the corresponding ground truth), and the translation is performed in latent space. Synthetic images and the real photos are first transformed to the same latent space with a shared variational autoencoder~\cite{kingma2013auto} (VAE). Meanwhile, another VAE is trained to project ground truth clean images into the corresponding latent space. The mapping between the two latent spaces is then learned with the synthetic image pairs, which restores the corrupted images to clean ones. The advantage of the latent restoration is that the learned latent restoration can generalize well to real photos because of the domain alignment within the first VAE. Besides, we differentiate the mixed degradation, and propose a partial nonlocal block that considers the long-range dependencies of latent features to specifically address the structured defects during the latent translation. In comparison with several leading restoration methods, we prove the effectiveness of our approach in restoring multiple degradations of real photos.

\section{Related Work}
\label{sec:relateG_work}
\noindent\textbf{Single degradation image restoration.} ~~
Existing image degradation can be roughly categorized into two groups: unstructured degration such as noise, blurriness, color fading, and low resolution, and structured degradation such as holes, scratches, and spots. For the former unstructured ones, traditional works often impose different image priors, including non-local self-similarity~\cite{buades2005non,mairal2009non,dabov2007image}, sparsity~\cite{elad2006image,mairal2007sparse,yang2010image,xie2012image} and local smoothness~\cite{weiss2007makes,babacan2008total,li2009markov}. Recently, a lot of deep learning based methods have also been proposed for different image degradation, like image denoising~\cite{zhang2017learning,zhang2017beyond,zhang2018ffdnet,mao2016image,lefkimmiatis2018universal,liu2018non,zhang2019rnan}, super-resolution~\cite{dong2014learning,kim2016accurate,ledig2017photo,wang2018esrgan,zhang2018residual}, and deblurring~\cite{xu2014deep,sun2015learning,nah2017deep,kupyn2018deblurgan}. 

Compared to unstructured degradation, structured degradation is more challenging and often modeled as the ``image painting" problem. Thanks to powerful semantic modeling ability, most existing best-performed inpainting methods are learning based. For example, \citet{liu2018image} masked out the hole regions within the convolution operator and enforces the network focus on non-hole features only.  To get better inpainting results, many other methods consider both local patch statistics and global structures. Specifically, \citet{yu2018generative} and \citet{liu2019coherent} proposed to employ an attention layer to utilize the remote context. And the appearance flow is explicitly estimated in \citet{ren2019structureflow} so that textures in the hole regions can be directly synthesized based on the corresponding patches.

No matter for unstructured or structured degradation, though the above learning-based methods can achieve remarkable results, they are all trained on the synthetic data. Therefore, their performance on the real dataset highly relies on synthetic data quality. For real old images, since they are often seriously degraded by a mixture of unknown degradation, the underlying degradation process is much more difficult to be accurately characterized. In other words, the network trained on synthetic data only, will suffer from the domain gap problem and perform badly on real old photos. In this paper, we model real old photo restoration as a new triplet domain translation problem and some new techniques are adopted to minimize the domain gap.

\vspace{0.5em}
\noindent\textbf{Mixed degradation image restoration.} ~~
In the real world, a corrupted image may suffer from complicated defects mixed with scratches, loss of resolution, color fading, and film noises. However, research solving mixed degradation is much less explored. The pioneer work~\cite{yu2018crafting} proposed a toolbox that comprises multiple light-weight networks, and each of them responsible for a specific degradation. Then they learn a controller that dynamically selects the operator from the toolbox. Inspired by~\cite{yu2018crafting}, \cite{suganuma2018attention} performs different convolutional operations in parallel and uses the attention mechanism to select the most suitable combination of operations. However, these methods still rely on supervised learning from synthetic data and hence cannot generalize to real photos. Besides, they only focus on unstructured defects and do not support structured defects like image inpainting. On the other hand, \citet{ulyanov2018deep} found that the deep neural network inherently resonates with low-level image statistics and thereby can be utilized as an image prior for blind image restoration without external training data. This method has the potential, though not claimed in~\cite{ulyanov2018deep}, to restore in-the-wild images corrupted by mixed factors. In comparison, our approach excels in both restoration performance and efficiency. 

\vspace{0.5em}
\noindent\textbf{Old photo restoration.}~~
Old photo restoration is a classical mixed degradation problem, but most existing methods ~\cite{stanco2003towards,bruni2004generalized,chang2005photo,giakoumis2005digital} focus on inpainting only. They follow a similar paradigm \ie, defects like scratches and blotches are first identified according to low-level features and then inpainted by borrowing the textures from the vicinity. However, the hand-crafted models and low-level features they used are difficult to detect and fix such defects well. Moreover, none of these methods consider restoring some unstructured defects such as color fading or low resolution together with inpainting. Thus photos still appear old fashioned after restoration. In this work, we reinvestigate this problem by virtue of a data-driven approach, which can restore images from multiple defects simultaneously and turn heavily-damaged old photos to modern style.

\section{Method}
In contrast to conventional image restoration tasks, old photo restoration is more challenging. First, old photos contain far more complex degradation that is hard to be modeled realistically and there always exists a domain gap between synthetic and real photos. As such, the network usually cannot generalize well to real photos by purely learning from synthetic data. Second, the defects of old photos is a compound of multiple degradations, thus essentially requiring different strategies for restoration. Unstructured defects such as film noise, blurriness and color fading, etc. can be restored with spatially homogeneous filters by making use of surrounding pixels within the local patch; structured defects such as scratches and blotches, on the other hand, should be inpainted by considering the global context to ensure the structural consistency. In the following, we propose solutions to address the aforementioned \emph{generalization issue} and \emph{mixed degradation issue} respectively.

\subsection{Restoration via latent space translation}
In order to mitigate the domain gap, we formulate the old photo restoration as an image translation problem, where we treat clean images and old photos as images from distinct domains and we wish to learn the mapping in between. However, as opposed to general image translation methods that bridge two different domains~\cite{isola2017image, CycleGAN}, we translate images across three domains: the real photo domain $\cR$, the synthetic domain $\cX$ where images suffer from artificial degradation, and the corresponding ground truth domain $\cY$ that comprises images without degradation. Such triplet domain translation is crucial in our task as it leverages the unlabeled real photos as well as a large amount of synthetic data associated with ground truth.

\begin{figure}[t!]
    \centering
    \small
    \begin{overpic}
        [scale=0.4]{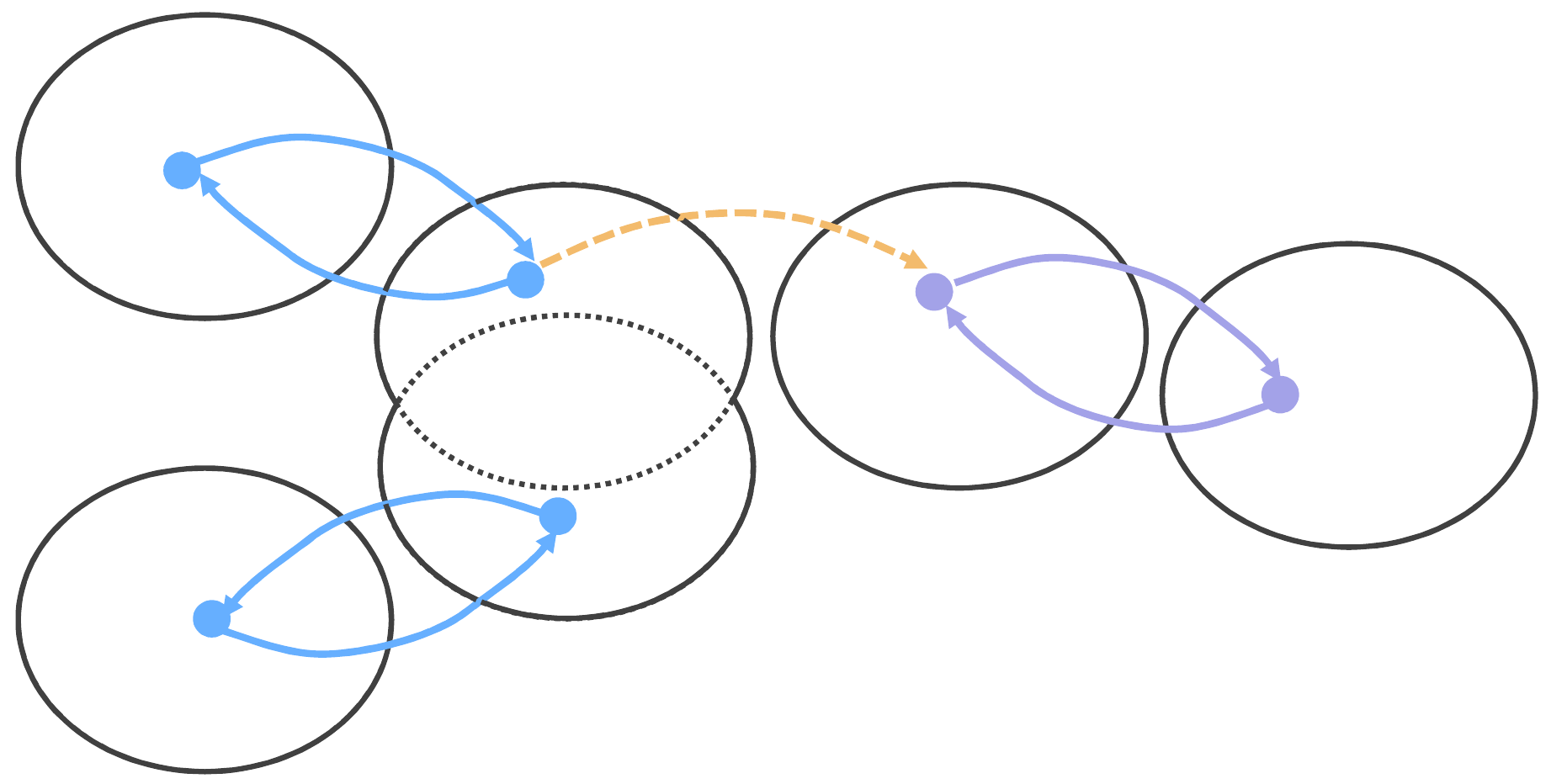} 
        \put(10,42){\footnotesize$x$} 
        \put(10,11.5){\footnotesize$r$}
        \put(84.5,25.5){\footnotesize$y$}
        \put(36,31){\footnotesize$z_x$}
        \put(37.5,15){\footnotesize$z_r$}
        \put(55.5,30){\footnotesize$z_y$}
        \put(11.5,50.5){\normalsize$\cX$}
        \put(11.5,21.5){\normalsize$\cR$}
        \put(34,40){\normalsize$\cZ_\cX$}
        \put(34,6.5){\normalsize$\cZ_{\cR}$}
        \put(60,40){\normalsize$\cZ_\cY$}
        \put(85,36.5){\normalsize$\cY$}
        \put(26,41){\footnotesize$E_{\cX}$}
        \put(18,19.5){\footnotesize$G_{\cR}$}
        \put(18,28.5){\footnotesize$G_{\cX}$}
        \put(25,5.5){\footnotesize$E_{\cR}$}
        \put(47,38){\footnotesize$T_{\cZ}$}
        \put(73,34){\footnotesize$G_{\cY}$}
        \put(70,19){\footnotesize$E_{\cY}$}
    \end{overpic}
    \caption{\textbf{Illustration of our translation method with three domains.} }
    \vspace{-1.5em}
    \label{fig:diagram1}
\end{figure}

We denote images from three domains respectively with  $r\in \cR$, $x\in \cX$ and $y\in \cY$, where $x$ and $y$ are paired by data synthesizing, \ie, $x$ is degraded from $y$. Directly learning the mapping from real photos $\{r\}_{i=1}^N$ to clean images $\{y\}_{i=1}^N$ is hard since they are not paired and thus unsuitable for supervised learning. We thereby propose to decompose the translation with two stages, which are illustrated in Figure~\ref{fig:diagram1}. First, we propose to map $\cR$, $\cX$, $\cY$ to corresponding latent spaces via $E_{\cR}:\cR \mapsto \cZ_{\cR}$, $E_{\cX}:\cX \mapsto \cZ_{\cX}$, and $E_{\cY}:\cY \mapsto \cZ_{\cY}$, respectively. In particular, because synthetic images and real old photos are both corrupted, sharing similar appearances, we align their latent space into the shared domain by enforcing some constraints. Therefore we have $\cZ_{\cR} \approx \cZ_{\cX}$. This aligned latent space encodes features for all the corrupted images, either synthetic or real ones. Then we propose to learn the image restoration in the latent space. Specifically, by utilizing the synthetic data pairs~$\{x,y\}_{i=1}^N$, we learn the translation from the latent space of corrupted images, $\cZ_{\cX}$, to the latent space of ground truth, $\cZ_{\cY}$, through the mapping $T_{\cZ}:\cZ_{\cX} \mapsto \cZ_{\cY}$, where $\cZ_{\cY}$ can be further reversed to $\cY$ through generator $G_{\cY}:{\cZ_{\cY}} \mapsto \cY$. By learning the latent space translation, real old photos $r$ can be restored by sequentially performing the mappings,
\vspace{-0.65em}
    \begin{equation}
        r_{\cR \to \cY} =  G_{\cY} \circ T_{\cZ} \circ E_{\cR}(r).
    \end{equation}

\begin{figure}[t!]
    \hspace{1em}
    \small
    \begin{overpic}
        [scale=0.58]{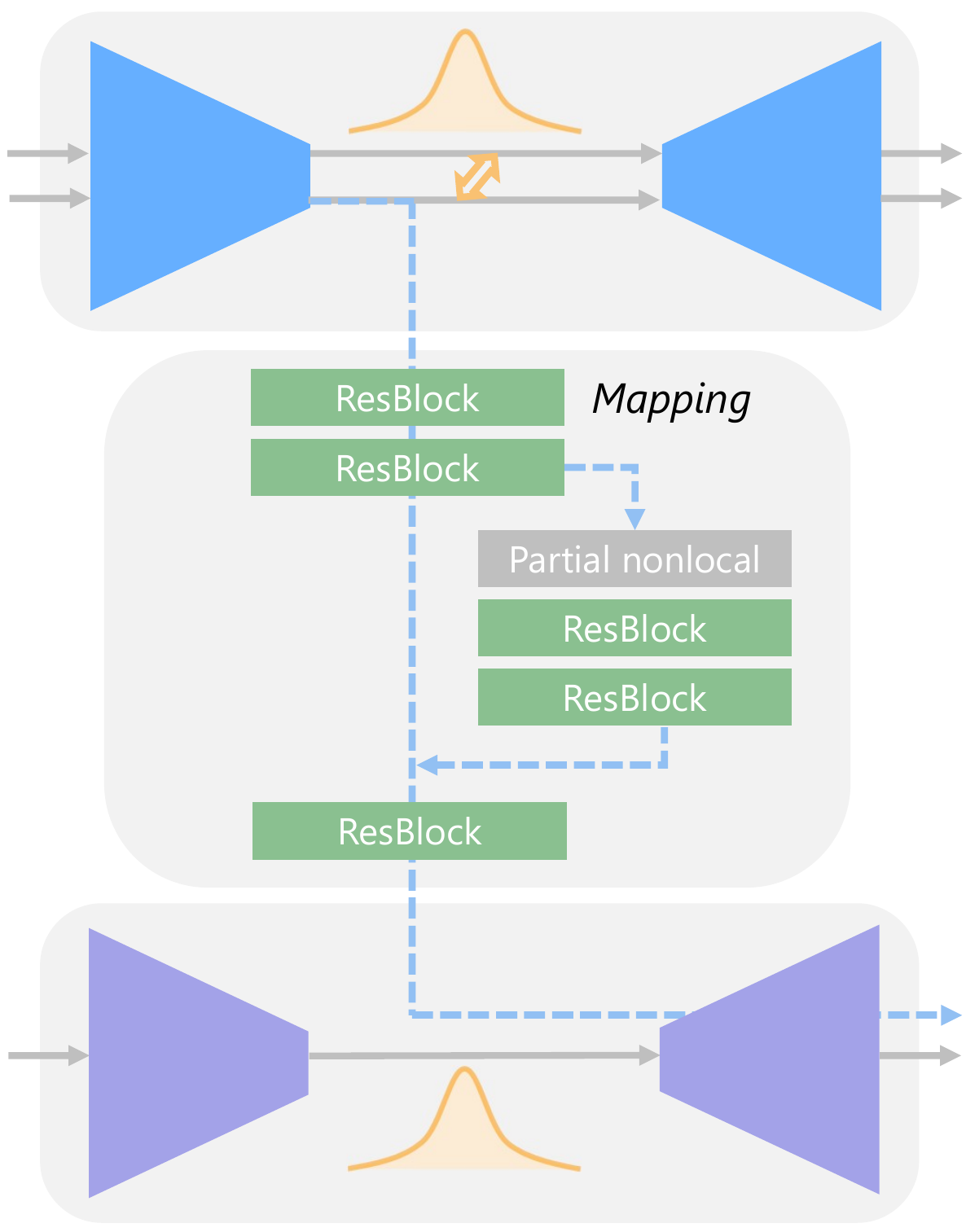} 
        \put(-2,86.5){$r$}
        \put(-2,83.5){$x$}
        \put(-2,14.5){$y$}
        \put(78,86.5){${r}_{\cR \to \cR}$}
        \put(78,83.5){${x}_{\cX \to \cX}$}
        \put(78,21){${r}_{\cR \to \cY}$}
        \put(78,17){${x}_{\cX \to \cY}$}
        \put(78,13){${y}_{\cY \to \cY}$}
        \put(11.5,84){\textcolor{white}{$E_{\cR,\cX}$}}
        \put(12.5,13){\textcolor{white}{$E_\cY$}}
        \put(59,84){\textcolor{white}{$G_{\cR,\cX}$}}
        \put(60.5,13){\textcolor{white}{$G_\cY$}}
        \put(35,80){$z_\cR, z_\cX$}
        \put(35,15.5){$z_\cY$}
        \put(35,19){$z_{\cR\to\cY},z_{\cX\to\cY}$}
        \put(42,93){$\cN(0,I)$}
        \put(42,9){$\cN(0,I)$}
        \put(42,84.5){$adv.$}
        \put(15,95.5){I.}
        \put(15,23){I.}
        \put(15,66){II.}
        \put(61.5,66.2){$\cT$}
    \end{overpic}
    \caption{\textbf{Architecture of our restoration network.} (I.) We first train two VAEs: VAE$_1$ for images in real photos $r\in\cR$ and synthetic images $x\in\cX$, with their domain gap closed by jointly training an adversarial discriminator; VAE$_2$ is trained for clean images $y\in\cY$. With VAEs, images are transformed to  compact latent space. (II.) Then,
    we learn the mapping that restores the corrupted images to clean ones in the latent space. }
    \vspace{-3em}
    \label{fig:diagram_new}
\end{figure}

\noindent\textbf{Domain alignment in the VAE latent space}~~ One key of our method is to meet the assumption that ${\cR}$ and ${\cX}$ are encoded into the same latent space. To this end, we propose to utilize variational autoencoder~\cite{kingma2013auto} (VAE) to encode images with compact representation, whose domain gap is further examined by an adversarial discriminator~\cite{GAN}. We use the network architecture shown in Figure~\ref{fig:diagram_new} to realize this concept.

In the first stage, two VAEs are learned for the latent representation. Old photos $\{r\}$ and synthetic images $\{x\}$ share the first one termed VAE$_1$, with the encoder $E_{\cR,\cX}$ and generator $G_{\cR,\cX}$, while the ground true images $\{y\}$ are fed into the second one, VAE$_2$ with the encoder-generator pair $\{E_{\cY},G_{\cY}\}$. VAE$_1$ is shared for both $r$ and $x$ in the aim that images from both corrupted domains can be mapped to a shared latent space. The VAEs assumes Gaussian prior for the distribution of latent codes, so that images can be reconstructed by sampling from the latent space. We use the re-parameterization trick to enable differentiable stochastic sampling~\cite{KingmaW13} and optimize VAE$_1$ with data $\{r\}$ and $\{x\}$ respectively. The objective with $\{r\}$ is defined as:
\begin{equation}
    \begin{split}
    \cL_{\text{VAE}_1}(r) &=\ {\mathrm{KL}}(E_{\cR,\cX}(z_r|r) || \cN({0},{I})) \\ &+ \alpha\Ebb_{z_r\sim E_{\cR,\cX}(z_r|r)}\left[\norm{ G_{\cR,\cX}(r_{\cR\to\cR}|z_r)-r}_1\right] \\&+ \cL_{\text{VAE}_1,\text{GAN}}(r)
\end{split}
\label{eq:vae1}
\end{equation}

where, $z_r\in\cZ_\cR$ is the latent codes for $r$, and $r_{\cR\to\cR}$ is the generation outputs. The first term in equations is the KL-divergence that penalizes deviation of the latent distribution from the Gaussian prior. The second $\ell_1$ term lets the VAE reconstruct the inputs, implicitly enforcing latent codes to capture the major information of images. Besides, we introduce the least-square loss (LSGAN)~\cite{mao2017least}, denoted as $\cL_{\text{VAE}_1,\text{GAN}}$ in the formula, to address the well-known over-smooth issue in VAEs, further encouraging VAE to reconstruct images with high realism. The objective with $\{x\}$, denoted as $\cL_{\text{VAE}_1}(x)$, is defined similarly. And VAE$_2$ for domain $\cY$ is trained with a similar loss so that the corresponding latent representation~$z_{y}\in \cY$ can be derived. 

We use VAE rather than vanilla autoencoder because VAE features denser latent representation due to the KL regularization (which will be proved in ablation study), and this helps produce closer latent space for $\{r\}$ and $\{x\}$ with VAE$_1$ thus leading to smaller domain gap. To further narrow the domain gap in this reduced space, we propose to use an adversarial network to examine the residual latent gap. Concretely, we train another discriminator $D_{\cR,\cX}$ that differentiates $\cZ_\cR$ and $\cZ_\cX$, whose loss is defined as,
\begin{equation}
    \begin{split}
    \cL_{\text{VAE}_1,\text{GAN}}^{\text{latent}}(r,x) =\ &\Ebb_{x\sim \cX}[D_{\cR,\cX}(E_{\cR,\cX}(x))^2] \\ &+ \Ebb_{r\sim \cR}[(1-D_{\cR,\cX}(E_{\cR,\cX}(r)))^2].
    \end{split}
\end{equation}
Meanwhile, the encoder $E_{\cR,\cX}$ of VAE$_1$ tries to fool the discriminator with a contradictory loss to ensure that $\cR$ and $\cX$ are mapped to the same space. Combined with the latent adversarial loss, the total objective function for VAE$_1$ becomes,
\begin{equation}
    \min_{E_{\cR,\cX},G_{\cR,\cX}}\max_{D_{\cR,\cX}}\ \cL_{\text{VAE}_1}(r) + \cL_{\text{VAE}_1}(x) + \cL_{\text{VAE}_1,\text{GAN}}^{\text{latent}}(r,x).
\end{equation}

\noindent\textbf{Restoration through latent mapping}~~
With the latent code captured by VAEs, in the second stage, we leverage the synthetic image pairs $\{x,y\}$ and propose to learn the image restoration by mapping their latent space (the mapping network $M$ in Figure~\ref{fig:diagram_new}). The benefit of latent restoration is threefold. First, as $\cR$ and $\cX$ are aligned into the same latent space, the mapping from $\cZ_{\cX}$ to $\cZ_{\cY}$ will also generalize well to restoring the images in $\cR$. Second, the mapping in a compact low-dimensional latent space is in principle much easier to learn than in the high-dimensional image space. In addition, since the two VAEs are trained independently and the reconstruction of the two streams would not be interfered with each other. The generator $G_{\cY}$ can always get an absolutely clean image without degradation given the latent code $z_{\cY}$ mapped from~$\cZ_{\cX}$, whereas degradations will likely remain if we learn the translation in pixel level.

Let $r_{\cR\to \cY}$, $x_{\cX\to \cY}$ and $y_{\cY \to \cY}$ be the final translation outputs for $r$, $x$ and $y$, respectively. At this stage, we solely train the parameters of the latent mapping network $\cT$ and fix the two VAEs. The loss function $\cL_{\cT}$, which is imposed at both the latent space and the end of generator $G_{\cY}$, consists of three terms,
\begin{equation}
    \cL_{\cT}(x,y) = \lambda_{1}\cL_{\cT,\ell_1} + \cL_{\cT,\text{GAN}} + \lambda_{2}\cL_{\text{FM}}
\label{eq:translation}
\end{equation}
where, the latent space loss, $\cL_{\cT,\ell_1} = \Ebb\norm{\cT(z_x)-z_y)}_1$, penalizes the $\ell_1$ distance of the corresponding latent codes. We introduce the adversarial loss $\cL_{\cT,\text{GAN}}$, still in the form of LSGAN~\cite{mao2017least}, to encourage the ultimate translated synthetic image $x_{\cX\to \cY}$ to look real. Besides, we introduce feature matching loss $L_{\text{FM}}$ to stabilize the GAN training. Specifically, $L_{\text{FM}}$ matches the multi-level activations of the adversarial network $D_{M}$, and that of the pretrained VGG network (also known as perceptual loss in~\cite{isola2017image,johnson2016perceptual}), \ie,
\begin{align}
    \mathcal{L}_{\text{FM}}
     =\ & \Ebb\ \Big[\sum_i \frac 1 {n_{D_{\cT}}^i}\| \phi_{D_{\cT}}^i (x_{\cX\to \cY}) - \phi_{D_{\cT}}^i (y_{\cY\to \cY})\|_1 \nonumber \\
    + & \sum_i \frac 1 {n_{\text{VGG}}^i} \| \phi_{\text{VGG}}^i (x_{\cX\to \cY}) - \phi_{\text{VGG}}^i (y_{\cY\to \cY})\|_1\Big],
\end{align}
where $\phi^i_{D_{\cT}}$ ($\phi^i_{\text{VGG}}$) denotes the $i^{th}$ layer feature map of the discriminator (VGG network), and $n^i_{D_{\cT}}$ ($n^i_{\text{VGG}}$)
indicates the number of activations in that layer.

\subsection{Multiple degradation restoration }
The latent restoration using the residual blocks, as described earlier, only concentrates on local features due to the limited receptive field of each layer. Nonetheless, the restoration of structured defects requires plausible inpainting, which has to consider long-range dependencies so as to ensure global structural consistency. Since legacy photos often contain mixed degradations, we have to design a restoration network that simultaneously supports the two mechanisms. Towards this goal, we propose to enhance the latent restoration network by incorporating a global branch as shown in Figure~\ref{fig:diagram_new}, which composes of a nonlocal block~\cite{wang2018non} that considers global context and several residual blocks in the following. While the original block proposed in~\cite{wang2018non} is unaware of the corruption area, our nonlocal block explicitly utilizes the mask input so that the pixels in the corrupted region will not be adopted for completing those area. Since the context considered is a part of the feature map, we refer to the module specifically designed for the latent inpainting as \emph{partial nonlocal block}.

Formally, let $F\in \Rbb^{C\times HW}$ be the intermediate feature map in $M$ ($C$, $H$ and $W$ are number of channels, height and width respectively), and $m\in \{0,1\}^{HW}$ represents the binary mask downscaled to the same size, where $1$ represents the defect regions to be inpainted and $0$ represents the intact regions. The affinity between $i^{th}$ location and $j^{th}$ location in $F$, denoted by $s_{i,j}\in\Rbb^{HW\times HW}$, is calculated by the correlation of $F_i$ and $F_j$ modulated by the mask $(1-m_j)$, \ie,
\begin{equation}
    s_{i,j}=(1-m_j)f_{i,j} / \sum_{\forall k} (1-m_k)f_{i,k},
\end{equation}
where,
\begin{equation}
    f_{i,j} = \exp(\theta(F_i)^T\cdot \phi(F_j))
\end{equation}
gives the pairwise affinity with embedded Gaussian. $\theta$ and $\phi$ project $F$ to Gaussian space for affinity calculation. According to the affinity $s_{i,j}$ that considers the holes in the mask, the partial nonlocal finally outputs
\begin{equation}
    O_i = {\nu}\left(\sum_{\forall j} s_{i,j} {\mu}(F_j)\right),
\end{equation}
which is a weighted average of correlated features for each position. We implement the embedding functions ${\theta}$, ${\phi}$, ${\mu}$ and ${\nu}$ with 1$\times$1 convolutions. 

 We design the global branch specifically for inpainting and hope the non-hole regions are left untouched, so we fuse the global branch with the local branch under the guidance of the mask, \ie,
\begin{equation}
    F_{fuse} = (1-m)\odot \rho_{\text{local}}(F) + m\odot \rho_{\text{global}}(O),
\end{equation}
where operator $\odot$ denotes Hadamard product, and $\rho_{\text{local}}$ and $\rho_{\text{global}}$ denote the nonlinear transformation of residual blocks in two branches. In this way, the two branches constitute the latent restoration network, which is capable to deal with multiple degradation in old photos. We will detail the derivation of the defect mask in Section~\ref{sec:implementation}.
 
\section{Experiment}
\subsection{Implementation}
\label{sec:implementation}
\noindent\textbf{Training Dataset}~~
We synthesize old photos using images from the Pascal VOC dataset~\cite{everingham2015pascal}. In order to render realistic defects, we also collect scratch and paper textures, which are further augmented with elastic distortions. We use layer addition, lighten-only and screen modes with random level of opacity to blend the scratch textures over the real images from the dataset. To simulate large-area photo damage, we generate holes with feathering and random shape where the underneath paper texture is unveiled. Finally, film grain noises and blurring with random amount are introduced to simulate the unstructured defects. Besides, we collect 5,718 old photos to form the images old photo dataset. 

\begin{figure}[t!]
    \begin{center}
    \includegraphics[width=0.7\linewidth]{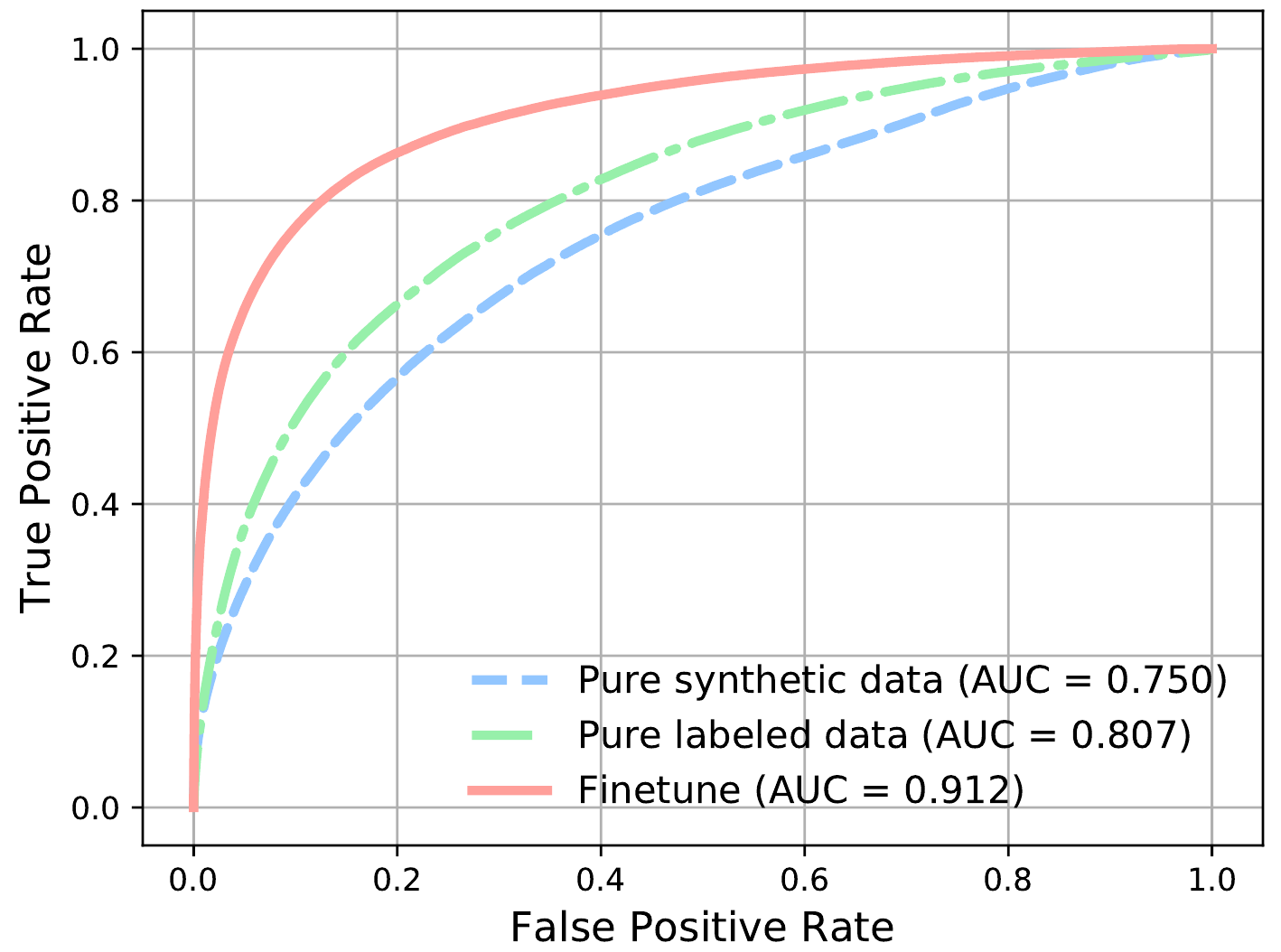}
    \end{center}
    \vspace{-1em}
    \caption{\textbf{ROC curve for scratch detection of different data settings.}}
    \label{figure:roc_curve}
    \vspace{-1.5em}
\end{figure}

\vspace{0.3em}
\noindent\textbf{Scratch detection}~~
To detect structured area for the parital nonlocal block, We train another network with Unet architecture~\cite{ronneberger2015u}. The detection network is first trained using the synthetic images only. We adopt the focal loss~\cite{lin2017focal} to remedy the imbalance of positive and negative detections. To further improve the detection performance on real old photos, we annotate 783 collected old photos with scratches, among which we use 400 images to finetune the detection network. The ROC curves on the validation set in Figure~\ref{figure:roc_curve} show the effectiveness of finetuning. The area under the curve (AUC) after finetuning reaches 0.91. 

\vspace{0.3em}
\noindent\textbf{Training details}~~
We adopt Adam solver~\cite{kingma2014adam} with $\beta_1=0.5$ and $\beta_2=0.999$. The learning rate is set to 0.0002 for the first 100 epochs, with linear decay to zero thereafter.  During training, we randomly crop images to 256$\times$256. In all the experiments, we empirically set the parameters in Equations~\eqref{eq:vae1} and \eqref{eq:translation} with $\alpha=10$, $\lambda_1=60$ and $\lambda_2=10$ respectively. 

\subsection{Comparisons}
\vspace{0.3em}
\noindent\textbf{Baselines}~~
We compare our method against state-of-the-art approaches. For fair comparison, we train all the methods with the same training dataset (Pascal VOC) and test them on the corrupted images synthesized from DIV2K dataset~\cite{Agustsson_2017_CVPR_Workshops} and the test set of our old photo dataset. The following methods are included for comparison.

\begin{itemize}[leftmargin=*]
\itemsep0em 
\item Operation-wise attention~\cite{suganuma2018attention} performs multiple operations in parallel and uses an attention mechanism to select the proper branch for mixed degradation restoration. It learns from synthetic image pairs with supervised learning. 
\item Deep image prior~\cite{ulyanov2018deep} learns the image restoration given a single degraded image, and has been proven powerful in denoising, super-resolution and blind inpainting.
\item Pix2Pix~\cite{pix2pixhd} is a supervised image translation method, which leverages synthetic image pairs to learn the translation in image level.
\item CycleGAN~\cite{CycleGAN} is a well-known unsupervised image translation method that learns the translation using unpaired images from distinct domains. 
\item The last baseline is to sequentially perform  BM3D~\cite{dabov2009bm3d}, a classical denoising method, and EdgeConnect~\cite{nazeri2019edgeconnect}, a state-of-the-art inpainting method, to restore the unstructured and structured defects respectively. 
\end{itemize}

\begin{figure*}[!t]
    \begin{center}
    \includegraphics[width=1\linewidth]{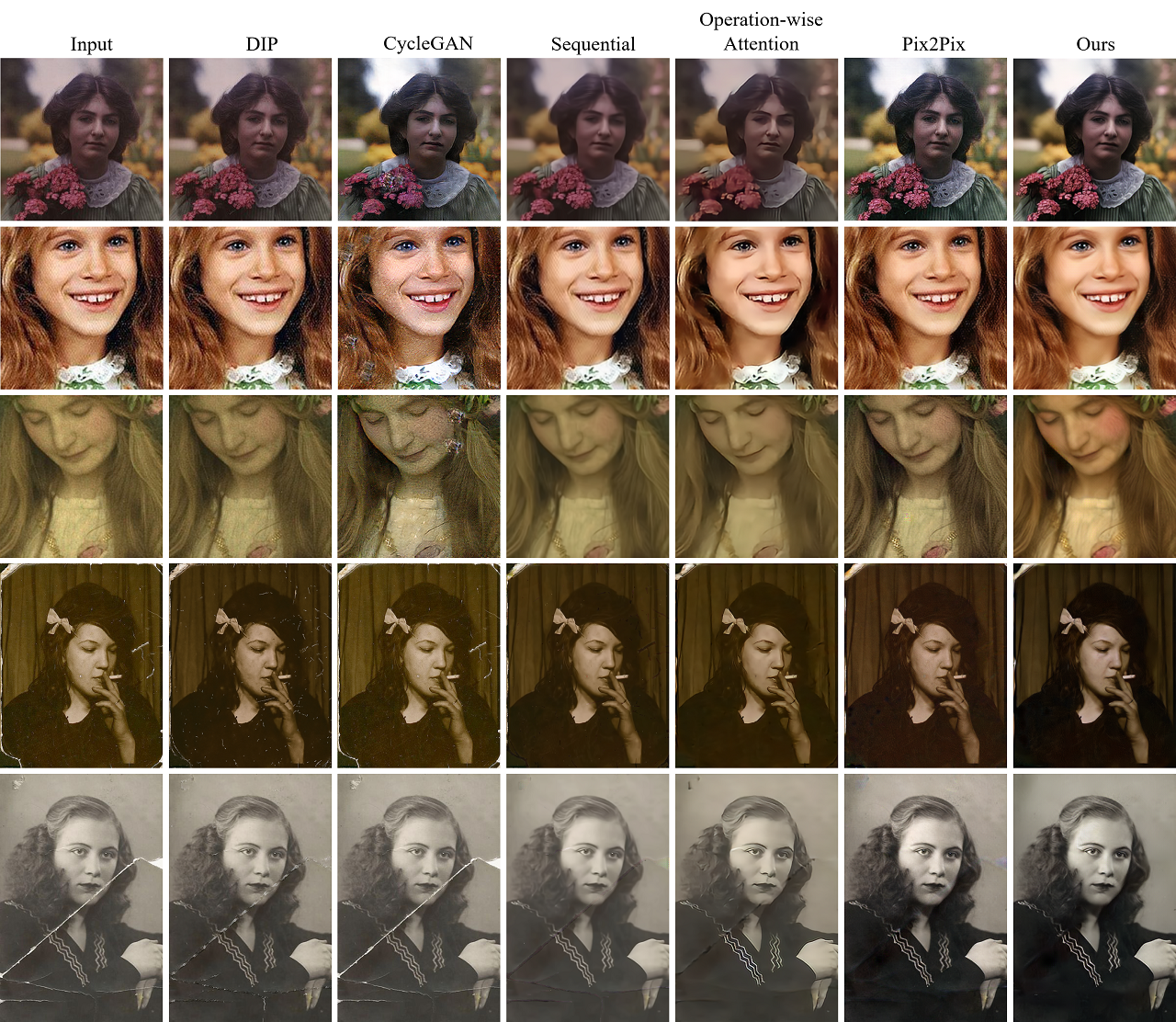}
    \end{center}
    \vspace{-0.8em}
    \caption{\textbf{Qualitative comparison against state-of-the-art methods.} It shows that our method can restore both unstructured and structured degradation and our recovered results are significantly better than other methods.}
    \label{figure:final_comparison}
    \vspace{-0.6em}
\end{figure*}

\begin{table}[t!]
    \small
    \begin{center}
    \setlength{\tabcolsep}{1.3mm}{
    \begin{tabular}{@{}lcccc@{}}
    \toprule
    Method& PSNR $\uparrow$& SSIM $\uparrow$ & LPIPS $\downarrow$ & FID $\downarrow$ \\ \midrule
    Input & 12.92& 0.49 & 0.59 & 306.80\\
    Attention~\cite{suganuma2018attention}& \textbf{24.12} & \textbf{0.70}& 0.33 & 208.11\\
    DIP~\cite{ulyanov2018deep} & 22.59& 0.57 & 0.54 & 194.55\\
    Pix2pix~\cite{pix2pixhd}& 22.18& 0.62 & \textbf{0.23}& \textbf{135.14}\\
    Sequential~\cite{dabov2009bm3d,nazeri2019edgeconnect} &22.71&0.60&0.49&191.98 \\
     \midrule
    Ours w/o PN& 23.14& 0.68 & 0.26 & {143.62}\\
    Ours w/ PN& \textbf{23.33}& \textbf{0.69} & \textbf{0.25}& \textbf{\textbf{134.35}}\\ \bottomrule
    \end{tabular}
    \caption{\textbf{Quantitative results on the DIV2K dataset.} Upward arrows indicate that a higher score denotes a good image quality. We highlight the best two scores for each measure. In the table, PN stands for partial nonlocal block.}
    \label{table:quantitative_ablation_partial_nl}
    }
    \vspace{-2em}
    \end{center}
\end{table}

\vspace{0.3em}
\noindent\textbf{Quantitative comparison}~~
We test different models on the synthetic images from DIV2K dataset and adopt four metrics for comparison. Table~\ref{table:quantitative_ablation_partial_nl} gives the quantitative results. The peak signal-to-noise ratio (PSNR) and the structural similarity index (SSIM) are used to compare the low-level differences between the restored output and the ground truth. The operational-wise attention method unsurprisingly achieves the best PSNR/SSIM score since this method directly optimizes the pixel-level $\ell_1$ loss. Our method ranks second-best in terms of PSNR/SSIM. However, these two metrics characterizing low-level discrepancy, usually do not correlate well with human judgment, especially for complex unknown distortions~\cite{zhang2018perceptual}. Therefore, we also adopt the recent learned perceptual image patch similarity (LPIPS)~\cite{zhang2018perceptual} metric which calculates the distance of multi-level activations of a pretrained network and is deemed to better correlate with human perception. This time, Pix2pix and our method give the best scores with a negligible difference. The operation-wise attention method, however, shows inferior performance under this metric, demonstrating it does not yield good perceptual quality. Besides, we adopt Fr\'echet Inception Distance (FID)~\cite{FID} which is widely used for evaluating the quality of generative models. Specifically, the FID score calculates the distance between the feature distributions of the final outputs and the real images. Still, our method and Pix2pix rank the best, while our method shows a slight quantitative advantage. In all, our method is comparable to the leading methods on synthetic data. 

\vspace{0.3em}
\noindent\textbf{Qualitative comparison}~~
To prove the generalization to real old photos, we conduct experiments on the real photo dataset. For a fair comparison, we retrain the CycleGAN to translate real photos to clean images. Since we lack the restoration ground truth for real photos, we cannot apply reference-based metrics for evaluation. Therefore, we qualitatively compare the results, which are shown in Figure~\ref{figure:final_comparison}. The DIP method can restore mixed degradations to some extent. However, there is a tradeoff between the defect restoration and the structural preservation: more defects reveal after a long training time while fewer iterations induce the loss of fine structures. CycleGAN, learned from unpaired images, tends to focus on restoring unstructured defects and neglect to restore all the scratch regions. Both the operation-wise attention method and the sequential operations give comparable visual quality. However, they cannot amend the defects that are not covered in the synthetic data, such as sepia issue and color fading. Besides, the structured defects still remain problematic, possibly because they cannot handle the old photo textures that are subtly different from the synthetic dataset. Pix2pix, which is comparable to our approach on synthetic images, however, is visually inferior to our method. Some film noises and structured defects still remain in the final output. This is due to the domain gap between synthetic images and real photos, which makes the method fail to generalize. In comparison, our method gives clean, sharp images with the scratches plausibly filled with unnoticeable artifacts. Besides successfully addressing the artifacts considered in data synthesis, our method can also enhance the photo color appropriately. In general, our method gives the most visually pleasant results and the photos after restoration appear like modern photographic images.

\begin{figure}[t!]
\begin{center}
\includegraphics[width=1.0\linewidth]{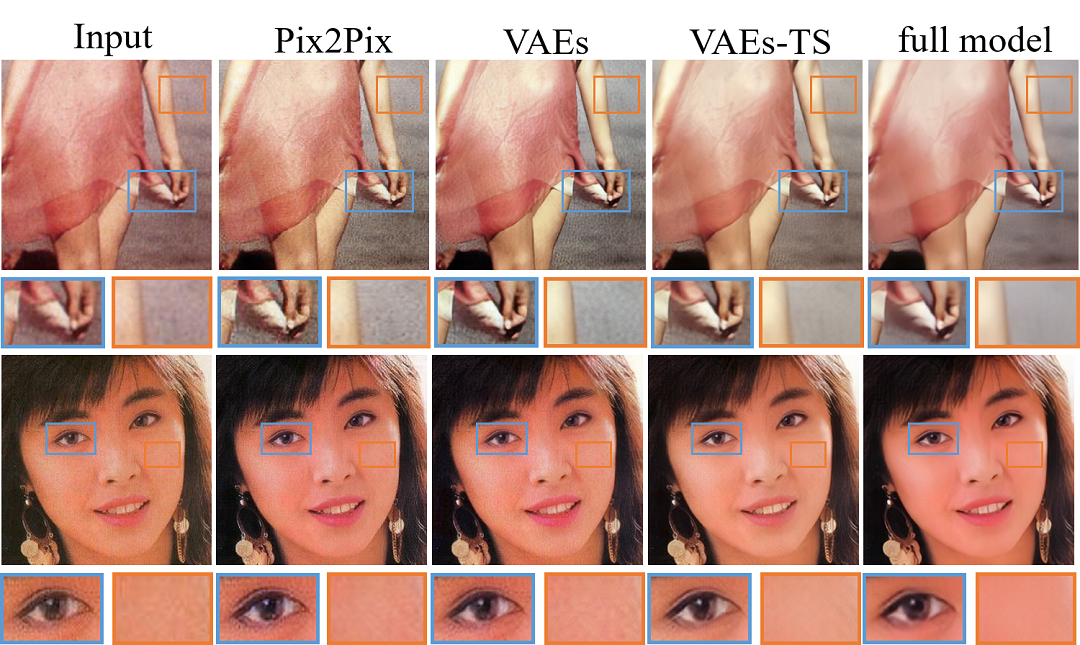}
\end{center}
\vspace{-1em}
\caption{\textbf{Ablation study for two-stage VAE translation.}}
\label{figure:qualitative_ablation_feature_translation}
\vspace{-0.6em}
\end{figure}

\begin{table}[t!]
    \small
    \begin{center}
    \setlength{\tabcolsep}{0.5mm}{
    \begin{tabularx}{\columnwidth}{@{}lYYYYY@{}}
\toprule
Method        & Top 1          & Top 2          & Top 3          & Top 4          & Top 5 \\ \midrule
DIP~\cite{ulyanov2018deep}           & 2.75           & 6.99           & 12.92          & 32.63          & 69.70 \\
CycleGAN~\cite{CycleGAN}      & 3.39           & 8.26           & 15.68          & 24.79          & 52.12 \\
Sequential~\cite{dabov2009bm3d,nazeri2019edgeconnect}     & 3.60           & 20.97          & 51.48          & 83.47          & 93.64 \\
Attention~\cite{suganuma2018attention}    & 11.22          & 28.18          & 56.99          & 75.85          & 89.19 \\
Pix2Pix~\cite{pix2pixhd}      & 14.19          & 54.24          & 72.25          & 86.86          & 96.61 \\ 
\textbf{Ours} & \textbf{64.83} & \textbf{81.35} & \textbf{90.68} & \textbf{96.40} & \textbf{98.72} \\ \bottomrule
    \end{tabularx}
    \caption{\textbf{User study results.} The percentage (\%) of user selection is shown.}
    \label{table:user_study}
    \label{table:quantitative_ablation_feature_translation}
    }
    \vspace{-2em}
    \end{center}
\end{table}

\noindent\textbf{User study}~~
To better illustrate the subjective quality, we conduct a user study to compare with other methods. We randomly select 25 old photos from the test set, and let users to sort the results according to the restoration quality. We collect subjective opinions from 22 users, with the results shown in Table~\ref{table:user_study}. It shows that our method is 64.86\% more likely to be chosen as the first rank result, which shows clear advantage of our approach.

\subsection{Ablation Study}
In order to prove the effectiveness of individual technical contributions, we perform the following ablation study. 

\noindent\textbf{Latent translation with VAEs}~~
Let us consider the following variants, with proposed components added one-by-one: 1) Pix2Pix which learns the translation in image-level; 2) two VAEs with an additional KL loss to penalize the latent space;  3) VAEs with two-stage training (VAEs-TS): the two VAEs are first trained separately and the latent mapping is learned thereafter with the two VAEs (not fixed); 4) our full model, which also adopts latent adversarial loss. We first calculate the Wasserstein distance~\cite{arjovsky2017wasserstein} between the latent space of old photos and synthetic images. Table~\ref{table:quantitative_ablation_feature_translation} shows that distribution distance gradually reduces after adding each component. This is because VAEs yield more compact latent space, the two-stage training isolates the two VAEs, and the latent adversarial loss further closes the domain gap. A smaller domain gap will improve the model generalization to real photo restoration. To verify this, we adopt a blind image quality assessment metric, BRISQUE~\cite{mittal2012no}, to measure photo quality after restoration. The BRISQUE score in Table~\ref{table:quantitative_ablation_feature_translation} progressively improves by applying the techniques, which is also consistent with the visual results in Figure~\ref{figure:qualitative_ablation_feature_translation}.

\begin{table}[t!]
    \small
    \begin{center}
    
    \setlength{\tabcolsep}{0.5mm}{
    \begin{tabularx}{\columnwidth}{@{}lYYYY@{}}
    \toprule
    Method  & Pix2Pix & VAEs &  VAEs-TS & full model\\ \midrule
    Wasserstein $\downarrow$& 1.837& 1.048 & 0.765  & \textbf{0.581}  \\ 
    BRISQUE $\downarrow$ & 25.549  & 23.949& 23.396 & \textbf{23.016} 
    \\ \bottomrule
    \end{tabularx}
    \caption{\textbf{Ablation study of latent translation with VAEs.}}
    \label{table:quantitative_ablation_feature_translation}
    }
    \end{center}
    \vspace{-1em}
\end{table}

\begin{figure}[t!]
    \begin{center}
    \includegraphics[width=1.0\linewidth]{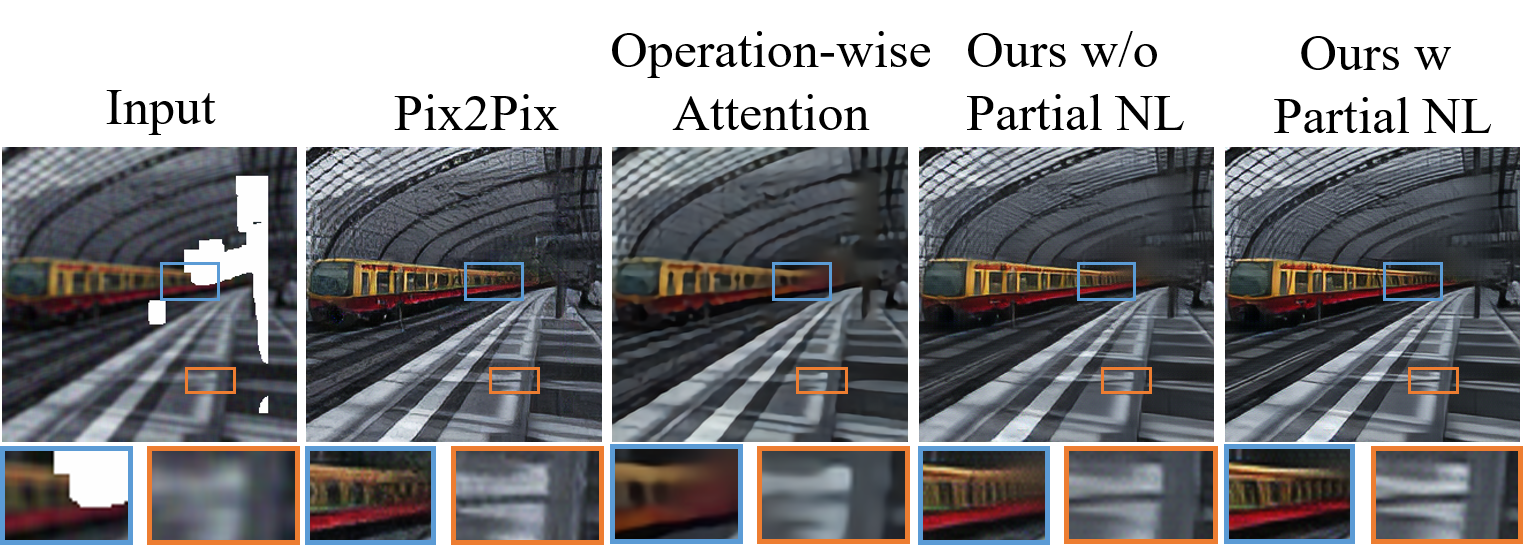}
    \end{center}
    \vspace{-1.4em}
    \caption{\textbf{Ablation study of partial nonlocal block.} Partial nonlocal better inpaints the structured defects.}
    \label{figure:qualitative_ablation_PN}
    \vspace{-0.8em}
\end{figure}

\begin{figure}[t!]
    \begin{center}
    \includegraphics[width=1.0\linewidth]{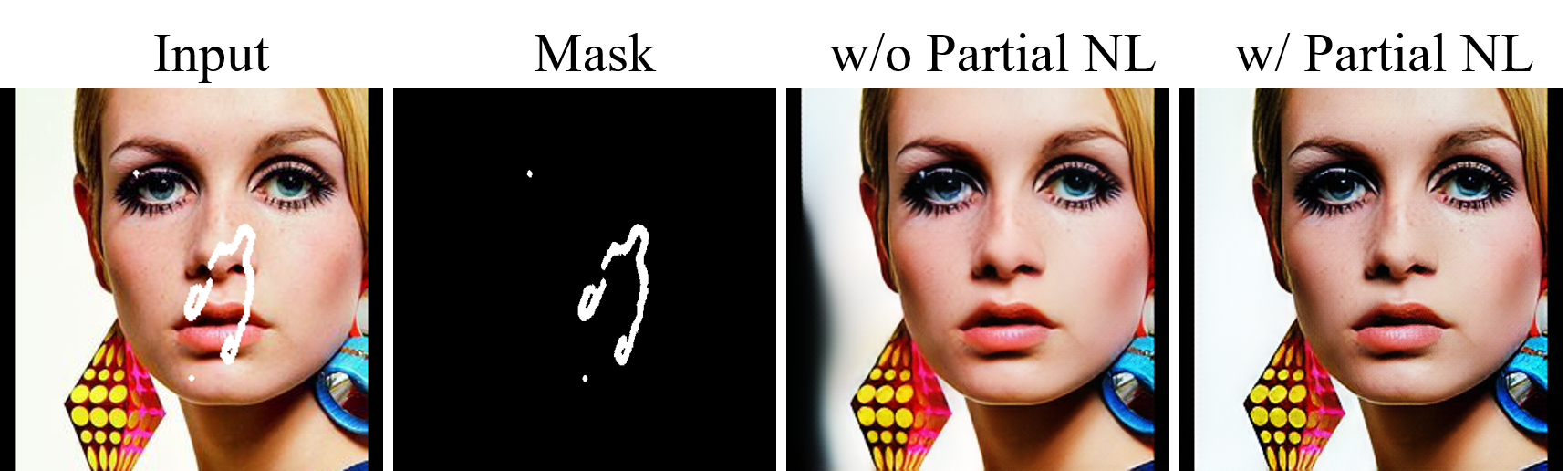}
    \end{center}
    \vspace{-1.4em}
    \caption{\textbf{Ablation study of partial nonlocal block.} Partial nonlocal does not touch the non-hole regions.}
    \label{figure:qualitative_ablation_PN_2}
    \vspace{-0.8em}
\end{figure}

\noindent\textbf{Partial nonlocal block}~~
The effect of partial nonlocal block is shown in Figure~\ref{figure:qualitative_ablation_PN} and~\ref{figure:qualitative_ablation_PN_2}. Because a large image context is utilized, the scratches can be inpainted with fewer visual artifacts and better globally consistent restoration can be achieved. Besides, the quantitative result in Table~\ref{table:quantitative_ablation_partial_nl} also shows that the partial nonlocal block consistently improves the restoration performance on all the metrics.

\setlength{\tabcolsep}{1.0pt}
\begin{figure}[t!]
\begin{center}
\includegraphics[width=1.0\linewidth]{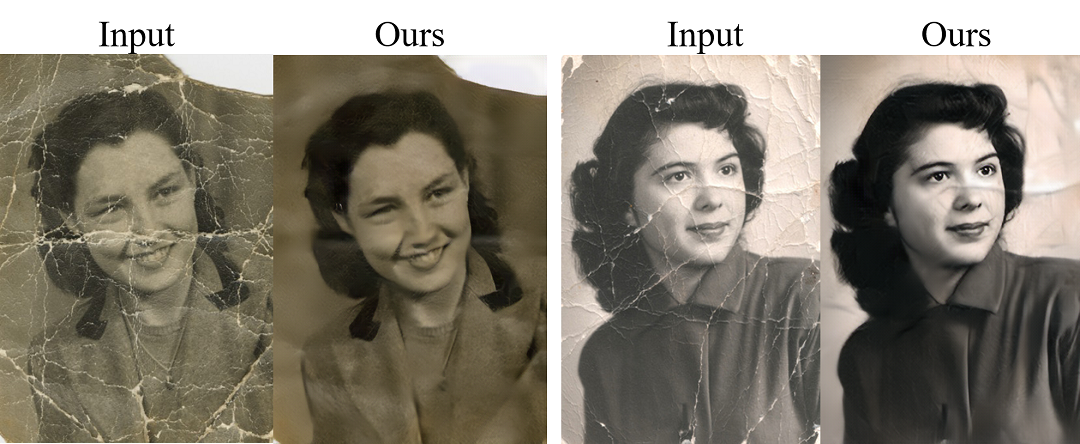}
\end{center}
\vspace{-1.4em}
\caption{\textbf{Limitation.} Our method cannot handle complex shading artifacts.}
\label{figure:limitation}
\vspace{-1em}
\end{figure}

\section{Discussion and Conclusion}
We propose a novel triplet domain translation network to restore the mixed degradation in old photos. The domain gap is reduced between old photos and synthetic images, and the translation to clean images is learned in latent space. Our method suffers less from generalization issue compared with prior methods. Furthermore, we propose a partial nonlocal block which restores the latent features by leveraging the global context, so the scratches can be inpainted with better structural consistency. Our method demonstrates good performance in restoring severe degraded old photos. However, our method cannot handle complex shading as shown in Figure~\ref{figure:limitation}. This is because our dataset contains few old photos with such defects. One could possibly address this limitation using our framework by explicitly considering the shading effects during synthesis or adding more such photos as training data.\\

\noindent\textbf{Acknowledgements: } We would like to thank Xiaokun Xie for his help and anonymous reviewers for their constructive comments. This work was partly supported by Hong Kong ECS grant No.21209119, Hong Kong UGC.

\cleardoublepage

\clearpage
{\small
\bibliographystyle{IEEEtranN}
\bibliography{restoration}
}

\end{document}